# Feature Selection for Gender Classification in TUIK Life Satisfaction Survey

A. ÇOBAN[1] and İ. TARIMER[2]

[1] Muğla Sıtkı Koçman University, Muğla/Turkey, adilcoban@mu.edu.tr
[2] Muğla Sıtkı Koçman University, Muğla/Turkey, itarimer@mu.edu.tr

*Abstract*— As known, attribute selection is a method that is used before the classification of data mining. In this study, a new data set has been created by using attributes expressing overall satisfaction in Turkey Statistical Institute (TSI) Life Satisfaction Survey dataset. Attributes are sorted by Ranking search method using attribute selection algorithms in a data mining application. These selected attributes were subjected to a classification test with Naive Bayes and Random Forest from machine learning algorithms. The feature selection algorithms are compared according to the number of attributes selected and the classification accuracy rates achievable with them. In this study, which is aimed at reducing the dataset volume, the best classification result comes up with 3 attributes selected by the Chi2 algorithm. The best classification rate was 73% with the Random Forest classification algorithm.

## I. INTRODUCTION

The concept of happiness is defined as "a state of prosperity from complete and continuous attainment of all wishes" [1]. Definitions of happiness based on philosophy and religion are used to mean a better life than emotion [2]. Corporations and public institutions want their customers and employees to be happy. Governments and societies want citizens to be happy also. They protect and evolve their assets in this regard.

New methods and approaches have been developed for how to do the most accurate measurement of economic, biological, religious, and psychological areas, along with increased work on the happiness measurement [3]. In this context, various scientific fields such as positive psychology and happiness economics have been derived [4] [5]. Until the measurement of happiness is measured, many subjects are included in the research [2].

Although the idea of happiness has been wondered by thinkers for thousands of years, it has only recently been possible to search and investigate systematically. A number of studies have been carried out since 1980 onwards [6], [7]. In addition to the concept of happiness, life satisfaction is considered as a more measurable concept. Happiness concepts representing lifelines, emotions and human consciousness have been used in [8] [9] [10].

Life satisfaction is a cognitive and judicial situation which expresses the evaluation of life as a whole. Happiness on the other hand is conceived as an emotional state produced by positive and negative events and experiences in the life of the individual. Although there are some correlations between happiness and life satisfaction at different levels, these concepts are still different [11].

The concept of subjective well-being, which we cannot separate from the concept of happiness, is defined as people's evaluations of their quality of life [6] [7]. Researches and surveys on life satisfaction and happiness have been used as subjective well-being indicators in [12].

Many internationally organizations collect data for life satisfaction surveys. The Nordic countries are the seven top ones amongst the best 10 countries in life satisfaction concept from globally scaled researches' reports. This result was reported in World Data Base of Happiness that archived by Erasmus University Rotterdam [13] and the World Happiness Report [14] prepared by the United Nations. Turkey has ranked 69[th] happiest country regarding to the World Happiness Report's research.

International happiness survey, organized by GALLUP Research Company, is also an example of global-scale research. According to the company's publication on International Happiness Day, the happiest countries are Latin American countries. These results, however, appear to be very different from those reported by the United Nations [15]. This difference suggests that the basic concept used in the two separate studies differs. The survey by the United Nations has focused on "life satisfaction" while GALLUP researches the concept of "happiness". According to research on positive experience made by GALLUP, Turkey took place in the bottom row [16].

There are institutions and organizations that conduct research on happiness in Turkey. Some of these are Zenna [17], Futurebright [18], Turkey Statistics Institute [19] and the Habitat Association [20]. Apart TSI, the organizations mentioned above offer services to the firms that want to develop marketing and sales strategies.

The first Life Satisfaction Survey made by

Turkey Statistical Institute (TSI), is conducted as an additional module to the Household Budget Survey in November 2003. Since 2004, the research has been conducted regularly every year and has been updated with different additions and developments until 2012. In 2013, the sample volume of the Life Satisfaction Survey was increased to give an estimate at provincial level in order to see the difference between provinces and regions and to create resources for other studies at province level. TSI has carried out the research mentioned above at the provincial level in 2013 and 2016 [21].

We focused on applying data mining techniques on TSI life satisfaction surveys and classifying the gender with variables of satisfaction in this study. For this purpose, a new data set was created with 21 attributes indicating satisfaction levels. The strength of the association of these attributes with the class was rated by attribute selection algorithms. Then the rated attributes were classified by Naive Bayes and Random Forest classification algorithms. Finally, feature selection algorithms were compared according to classification accuracy ratios.

## II. MATERIAL AND METHOD

In this study, Orange Data Mining program [22] is preferred to use. The Orange program is an open source software package that can also be run as a visual programming tool or as a script in the Python programming language platform. In the orange library, data is filtered with hierarchically arranged components such as probability assessment, attribute rating and classification [23].

The Rank widget scores the attributes according to their correlation with the class. Attribute scoring methods that can be used in Rank widget are Information Gain, Information Gain Ratio, Gini Impurity Index, Chi-Square-Chi-Square, ReliefF, Fast Correlation Based Filter (FCBF) [24].

The feature selection algorithms used in the study are briefly introduced:

*Information Gain*: The ID3 algorithm uses information gain when constructing decision trees. When choosing an attribute, it is used to reduce the variance of multi-valued attributes taking into account the number and size of branches. It shows how close we are to the classification results given by the sub-values of the tested object. The concept of information acquisition is explained by the concept of entropy. Entropy is a concept that expresses irregularity. The lower the entropy value in the nodes of decision tree, the better the distinction is made. The entropy gets value between 0 (lowest value) and 1 (highest value) [25] [26].

*Information Gain Ratio*: An improved version of the Information Gain algorithm is information gain ratio. It is used because it solves the disadvantage of Information Gain. Although Information Gain is often a good measure to determine the degree of importance of qualification, a considerable bias problem arises. Information Gain Ratio is used to overcome this problem [27].

*Gini Impurity Index*: Used by the CART algorithm. A decision tree finds the attributes to be used when branching the structure. Attributes that can make the most discrimination are preferred because they get higher scores. If there is only one class label in the branch, no new branching occurs. As in the concept of entropy, the lower the purity index of the nodes is, the better the discrimination is made [28].

*Chi2* (Chi-square distribution): A statistical tool used for rejecting derived hypotheses that data are independent. It is a non-parametric technique used to determine whether the distribution of observed frequencies differs from the expected theoretical frequencies [29].

*ReliefF*: This is an algorithm which is not dependent on heuristic methods but works on low-order polynomial time, and is based on the feature selection algorithm which is resistant to noise and feature interactions and is used in binary classification [30].

*Fast Correlation-Based Filter (FCBF)*: It has been developed with the enhancement of the correlation-based (CBF) algorithm. Without a bidirectional correlation analysis, it is a fast algorithm that can determine related attributes and the redundancies between them [31].

Within the scope of the study, 196,203 records were used which are all the raw data of TSI Life Satisfaction Survey 2013. 83074 of them are male and 113129 of them are female. Expressing general satisfaction attributes and gender attribute from the data were selected to create a new data set through Microsoft Excel 2016. Descriptions of attributes are given in Table 1. Figure 1 shows the steps followed in the experimental works.

Table 1 Questions and Variables

|    | Attributes | Questions |
|----|------------|-----------|
| 1  | Personal Health | B12.1 Are you satisfied with your health? |
| 2  | Marriage | B12.2 Are you satisfied with your marriage? |
| 3  | Personal Education | B12.3 Are you satisfied with education you have received so far? |
| 4  | Housing | B12.4 Are you satisfied with the housing you are in? |
| 5  | District | B12.5 Are you satisfied with your neighborhood or district? |
| 6  | Job | B12.6 Are you satisfied with your job? |
| 7  | Job Income | B12.7 Are you satisfied with your income earned from your job? |
| 8  | Household Income | B12.8 Are you satisfied with your monthly household income? |
| 9  | Social life | B12.9 Are you satisfied with your social life (such as entertainment, cultural and sports)? |
| 10 | Self-care | B12.10 Are you satisfied with the time set aside for yourself? |
| 11 | Time spent on traffic to and from work | B12.11 Are you satisfied with the time you spend in traffic to go to work? |
| 12 | Relative | B13.1 Are you satisfied with your relationships with your relatives? |
| 13 | Friend | B13.2 Are you satisfied with your relationships with your friends? |
| 14 | Neighbor | B13.3 Are you satisfied with your relationships with your neighbors? |
| 15 | Workplace Relations | B13.4 Are you satisfied with relationships with people related to your work? |
| 16 | General Health Services | B14.1 Are you satisfied with your health services? |
| 17 | Public order | B14.2 Are you satisfied with public security services? |
| 18 | Judicial | B14.3 Are you satisfied with judicial services? |
| 19 | General Education | B14.4 Are you satisfied with the education services? |
| 20 | SII Services | B14.5 Are you satisfied with Social Insurance Institution Services? |
| 21 | Transportation | B14.6 Are you satisfied with the transportation services? |

Answers to the satisfaction scale questions are evaluated by five scores:
*1 – Strongly satisfied     2 – Satisfied     3 – Undecided     4 – Disagreed     5 – Strongly disagree*

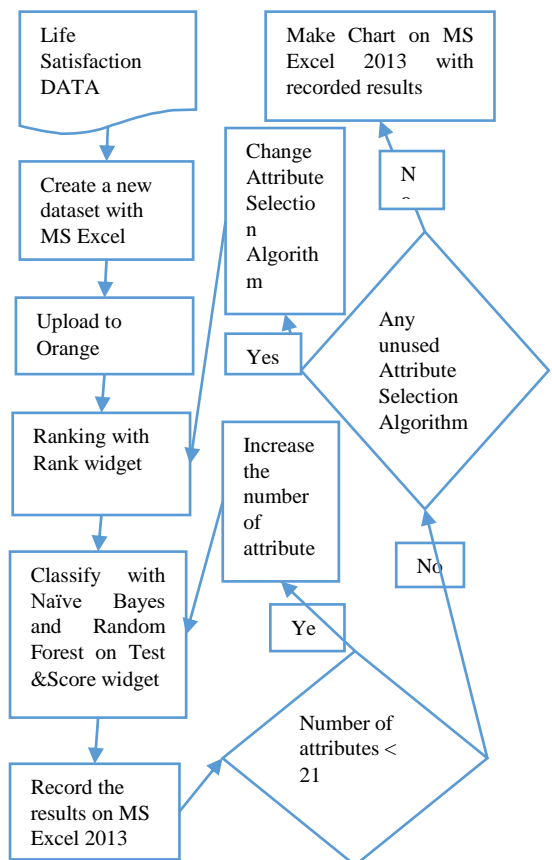

Figure 1: Experiment Steps

Figure 1 shows the experiment steps proceeded in the study. The data cleared through Microsoft Excel 2013 was uploaded to Orange with the File widget in the data mining program. The class attribute is selected as target in the interface of the File widget and selections are made for each attribute to evaluate them as nominal. The attributes were scored attribute selection algorithm in the Rank widget. Once ranked according to their score in the Rank interface, two of attributes selected to send Test and Score widget. The number of attributes was increased in each round and sent to the Test and Score widget. In Rank widget, each ranking method was performed separately and the classification test was repeated. In the Test and Score interface, Cross Validation was selected as the sampling method and 10 folds were used. The classification methods used during the test and score phase were Naive Bayes and Random Forest. The accuracy ratios obtained after each test were transferred to in a MS Excel table and visualized by charts. The data processing model generated in the Orange data mining program interface of this model is shown in Fig 2. The interfaces of the widgets used in the data processing model are shown in Fig 3, 4 and 5.

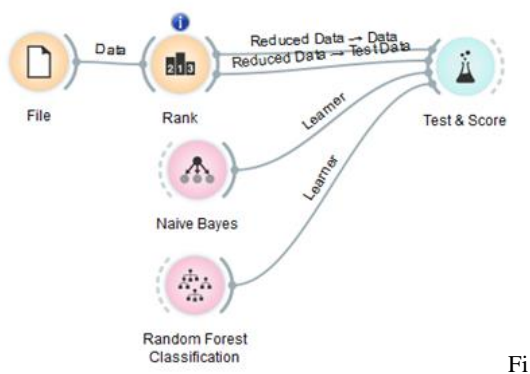

Figure 2: Model used on Orange Data Mining

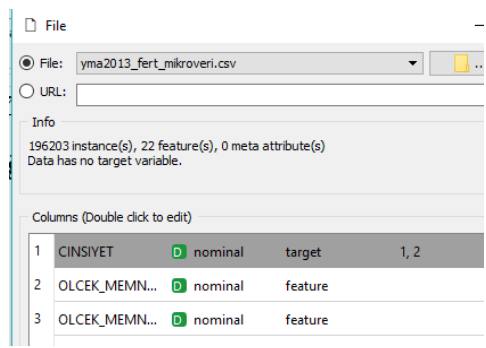

Figure 3: File

In Figure 3 shows that selecting the file to be worked on, determination of the type of attributes (nominal) and determination of target attribute are done.

Figure 4: Rank

In Figure 4, the scored attributes can be seen as ranked.

Figure 5: Test and Score

In Figure 5, classification test result is shown.

III. FINDINGS AND RESULTS

In this section, the results produced from the experiments in the study were obtained and these results are converted into tables and charts to be compared with each other. In this context; attributes according to result obtained from the feature selection algorithms; the number is increased every time and subjected to classification test for each algorithm separately. The scores obtained by attribute scoring methods are shown in Table 2.

Table 2 Scores by Attribute Selection Algorithm

| No. | Attributes | Attribute Selection Algorithms | | | | | |
|---|---|---|---|---|---|---|---|
| | | Inf. gain | Gain Ratio | Gini | Chi2 | ReliefF | FCBF |
| 3 | Personal Education | **4,523E-02** | **2,245E-02** | **2,815E-02** | **1,101E+04** | **5,000E-02** | **3,157E-02** |
| 18 | Judicial | **2,233E-02** | **1,272E-02** | **1,493E-02** | **4,393E+03** | 1,000E-02 | **1,653E-06** |
| 20 | SII Services | **1,956E-02** | **1,218E-02** | **1,271E-02** | 2,894E+03 | -2,200E-02 | 1,553E-06 |
| 1 | Personal Health | 1,522E-02 | 9,483E-03 | 1,010E-02 | 2,148E+03 | -1,000E-02 | 1,212E-06 |
| 7 | Job income | 1,158E-02 | 5,767E-03 | 7,261E-03 | **3,848E+03** | 9,425E-04 | 1,900E-07 |
| 2 | Marriage | 7,142E-03 | 7,001E-03 | 4,598E-03 | 3,988E+02 | 1,988E-02 | 1,233E-07 |
| 17 | Public order | 5,672E-03 | 4,854E-03 | 3,882E-03 | 2,708E+02 | -4,000E-02 | 5,263E-07 |
| 21 | Transportation | 5,630E-03 | 4,032E-03 | 3,677E-03 | 1,938E+02 | **3,400E-02** | 4,850E-07 |
| 4 | Housing | 2,755E-03 | 2,127E-03 | 1,823E-03 | 3,518E+02 | **4,000E-02** | 2,487E-07 |
| 19 | General Education | 2,692E-03 | 1,854E-03 | 1,836E-03 | 1,287E+02 | -4,800E-02 | 2,227E-07 |
| 16 | General Health Services | 2,254E-03 | 1,588E-03 | 1,541E-03 | 1,995E+00 | 3,197E-17 | 1,888E-07 |
| 12 | Relative | 2,024E-03 | 1,801E-03 | 1,383E-03 | 1,533E+02 | 1,600E-02 | 1,930E-07 |
| 14 | Neighbor | 9,777E-04 | 9,081E-04 | 6,665E-04 | 6,314E+01 | -3,200E-02 | 9,578E-08 |
| 10 | Self-care | 8,347E-04 | 5,185E-04 | 5,697E-04 | 3,313E+00 | -1,600E-02 | 6,502E-08 |
| 9 | Social Life | 6,866E-04 | 3,914E-04 | 4,692E-04 | 2,790E+01 | -2,000E-03 | 5,067E-08 |
| 8 | Household Income | 6,678E-04 | 3,497E-04 | 4,532E-04 | 6,847E+00 | 2,000E-03 | 4,692E-08 |
| 13 | Friend | 5,658E-04 | 6,102E-04 | 3,858E-04 | 9,185E+00 | -4,000E-03 | 5,972E-08 |
| 5 | District | 3,728E-04 | 3,316E-04 | 2,498E-04 | 2,662E+01 | -2,000E-03 | 3,610E-08 |
| 11 | Time spent on traffic to and from work | 2,071E-04 | 1,372E-04 | 1,199E-04 | 1,142E+03 | 2,171E-03 | 3,698E-07 |
| 6 | Job | 1,785E-04 | 1,334E-04 | 9,802E-05 | 5,948E+02 | 1,794E-02 | 4,410E-07 |
| 15 | Workplace Relations | 6,071E-05 | 6,179E-05 | 3,474E-05 | 2,803E+01 | 8,533E-03 | **6,046E-03** |
| | | *The numbers marked in bold are the highest three values in the sequence for each algorithm. | | | | | |

The results shown in Table 2 demonstrates that; education, judicial and SII services attributes get the highest score by three of the algorithms (Inf. Gain, Gain Ratio and Gini). The Chi2 algorithm gives the highest score in terms of the income attribute in addition to the training and forensic attributes. The ReliefF algorithm gives the highest score to residential attribute besides education attribute and transportation attribute. FCBF, on the other hand, gives the highest scores to education attribute, judicial and business relation attributes.

In Figure 6, Naive Bayes classification test results are shown by a chart.

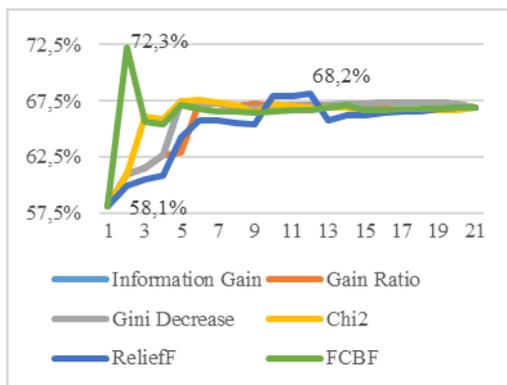

Figure 6 Naive Bayes classification results

From Figure 6, the highest value of the feature selection algorithms tested with Naive Bayes is seen as 72.3%. This value was obtained by selecting two attributes (3 and 15) with the highest score selected by FCBF algorithm. Information gain, Gini Decrease and Chi2 were obtained with the highest score of 67.5 with five attributes. The highest score (67.4) obtained by Gain Ratio was reached with six attributes. The highest score of ReliefF, 68.2, was reached with twelve attributes. As the number of attributes is increased, the scores obtained are retained same (no increase). The classification score value does not increase further because the contribution scores of the subsequently added attributes are low.

In Figure 7, test results of the Random Forest classification are shown by a chart.

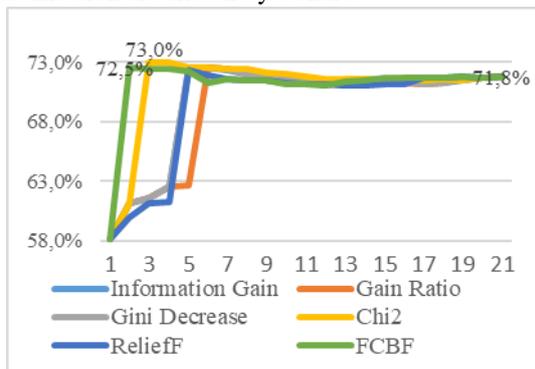

Figure 7 Random Forest classification results

As shown in Fig. 7, the highest value of the feature selection algorithms tested with Random Forest is seen as 73.0%. This was obtained when 3 attributes (3, 18, and 7) with the highest score were selected by the Chi2 selection algorithm. The highest score of 72.6 obtained with information gain and gini decrease was reached with five attributes. The highest score of 72.5 was reached with six attributes selected by Gain Ratio. The highest score of ReliefF, 72.2, was reached with five attributes. The highest score of 72.5, was reached with two attributes (3 and 15) selected by FCBF algorithm.

The highest classification score was obtained with personal training, judicial and job income attributes. The best result has been achieved using only these three attributes without having to use other attributes.

IV. DISCUSSION AND CONCLUSIONS

In this study, it is stated that data mining techniques can be applied on life satisfaction questionnaires and gender classification can be done with satisfaction variables. Therefore, at this point, it has been determined whether there is a difference of satisfaction among the genders. The accuracy of 73% in the classification confirms that the satisfaction levels differ between men and women for different attributes.

The satisfaction of personal education was selected among the first three most valuable attributes by all algorithms. Judicial satisfaction was also selected among the first three most valuable attributes by five algorithms. SII Services was also selected as previous mentioned attributes by four algorithms. Since these attributes are good at distinguishing gender, it has been seen that men and women participants have different thoughts in these matters. This suggests that the services received at these topics are assessed differently by men and women participants.

Another consequence of this work is that it is difficult to apply data mining methods to datasets that have large volume of data. In such a case, it is necessary to decrease variables by subsetting most valuable attributes, so irrelevant features are dismissed and classification accuracy is increased.

The disadvantages (class imbalance and subjectivity in the answers to the questions) in this dataset obtained from the life satisfaction survey were confronted as a problem when analyzing the dataset. For this reason, if the data set we have studied is also handled with algorithms developed for imbalanced data, higher accuracy percentages can be achieved.

V. ACKNOWLEDGEMENTS

The authors of this work would like to thank that Turkey Statistical Institute have given the life satisfaction data us to use in our thesis.